\documentclass[twoside,compsoc]{IEEEtran}
\usepackage{makecell}
\usepackage{hyperref}
\usepackage{array}
\usepackage{graphicx,amssymb,amsmath}
\usepackage{multicol}
\usepackage[noadjust]{cite}
\usepackage{setspace}
\usepackage{subfigure}
\usepackage{float}
\usepackage{url}
\usepackage[switch, pagewise]{lineno}
\usepackage{stfloats}
\usepackage{amsthm,pifont}
\usepackage{flushend}
\usepackage{cases,subeqnarray}
\usepackage{bm, multirow,bigstrut}
\usepackage{amsmath, amsthm, amssymb}
\usepackage{textcomp}
\usepackage{latexsym,bm}
\usepackage{booktabs}
\usepackage{xcolor}
\usepackage{mathtools}
\usepackage{dsfont}
\usepackage{extarrows}
\usepackage{epsfig}
\usepackage{epsfig}
\usepackage{epstopdf}
\usepackage{colortbl}
\usepackage[noend]{algpseudocode}
\usepackage{algorithmicx,algorithm}
\usepackage{threeparttable}
\IEEEoverridecommandlockouts
\begin{document}
\title{Experience Scaling: Post-Deployment Evolution For Large Language Models}
\author{Xingkun Yin, Kaibin Huang, Dong In Kim, Hongyang Du
\thanks{X. Yin, K. Huang, H. Du are with the Department of Electrical and Electronic Engineering, University of Hong Kong, Pok Fu Lam, Hong Kong SAR, China (Email: yinxingkun@connect.hku.hk, huangkb@hku.hk, duhy@hku.hk).
D. I. Kim is with the Department of Electrical and Computer Engineering, Sungkyunkwan University, Korea (Email: dongin@skku.edu).}
}
\maketitle
\vspace{-1cm}
\begin{abstract}
Scaling model size, training data, and compute power have driven advances in large language models (LLMs), but these approaches are reaching saturation as human-generated text is exhausted and further gains diminish. We propose experience scaling, a framework for continuous post-deployment evolution for LLMs through autonomous interaction with the environment and collaborative sharing of accumulated experience. The framework captures raw interactions, distills them into compact, reusable knowledge, and periodically refines stored content to preserve relevance and efficiency. We validate the framework in simulated real-world scenarios involving generalization to previously unseen but related tasks, repetitive queries, and over-saturated knowledge stores. Across all settings, experience scaling improves accuracy, sustains performance over time, and maintains gains when applied to novel situations. These results demonstrate that structured post-deployment learning can extend LLM capabilities beyond the limits of static human-generated data, offering a scalable path for continued intelligence progress.
\end{abstract}
\IEEEpeerreviewmaketitle

\section{Introduction}\label{sec_introduction}
Large language models (LLMs) have become a foundation for a wide range of applications~\cite{10.1145/3735632, Argyle2025ArtiFickleIntelligence}, from natural language interfaces to scientific analysis. The rapid adoption of the transformer architecture~\cite{vaswani2023attentionneed, kaddour2023challengesapplicationslargelanguage} was enabled by a consistent rise in model capability, which has followed a predictable pattern known as the scaling law. First formalized in 2020~\cite{kaplan2020scalinglawsneurallanguage}, this empirical principle shows performance improves systematically with increases in model parameters, training data, and compute power. Three scaling methods were developed under this guideline. Pre-training scaling expands models and datasets to capture broad linguistic patterns and world knowledge~\cite{radford2018improving}. Post-training scaling requires high-quality optimized data to align LLMs outputs with human preferences better~\cite{Zhao2024LRQuantLA}. Test-time scaling~\cite{NEURIPS2024_8b970e15} allocates additional computational resources during inference, typically by generating long chain-of-thought reasoning paths~\cite{wei2022chain}, to enhance results on complex and challenging tasks. These strategies have defined the dominant pathway for progress in LLMs, enabling significant performance gains across multiple domains.

However, while continuing to deliver LLMs that are more powerful than earlier generations in scale and applied benchmarks, the growth of their fundamental capabilities has slowed, indicating a plateau in improvement under conventional scaling paradigms~\cite{Ito2024QuantifyingAI}. One of the main constraints is the exhaustion of high-quality human-generated data~\cite{villalobos2024rundatalimitsllm,Xu_2025}. Leading LLMs have been reported to consume most publicly accessible digital and physical data available, and their rate of data usage has exceeded the pace at which humans can generate new data~\cite{Giataganas_2022}. While using synthetic data, as in the training of models like Llama-2~\cite{touvron2023llama2openfoundation}, Llama 3~\cite{grattafiori2024llama3herdmodels} and GPT-4~\cite{openai2024gpt4technicalreport}, helped alleviate part of the data shortage, it raises risks such as reduced diversity, reinforcement of existing biases, and overfitting to self-generated patterns.

One way to overcome the current data shortage is to reduce LLMs scalings' dependence on human-created content~\cite{villalobos2024rundatalimitsllm}. Human population size inherently constrains the rate of new data production, while the number of deployed LLMs is exponentially growing, and their potential perceptual channels are far broader than those available to humans~\cite{lin2025pushinglargelanguagemodels}. Specifically, human observation of the world is biologically limited, e.g., via language, sight, and hearing, while LLMs can absorb information in various formats from diverse sources at far greater temporal and spatial resolution~\cite{funke2021pointscheckcomparingvisual}. Yet, the vast amount of data generated from LLMs' interactions with the world remains largely unused in model development~\cite{NEURIPS2023_1b44b878}.

\begin{figure*}[h]
\centering
\includegraphics[width = 0.9\textwidth]{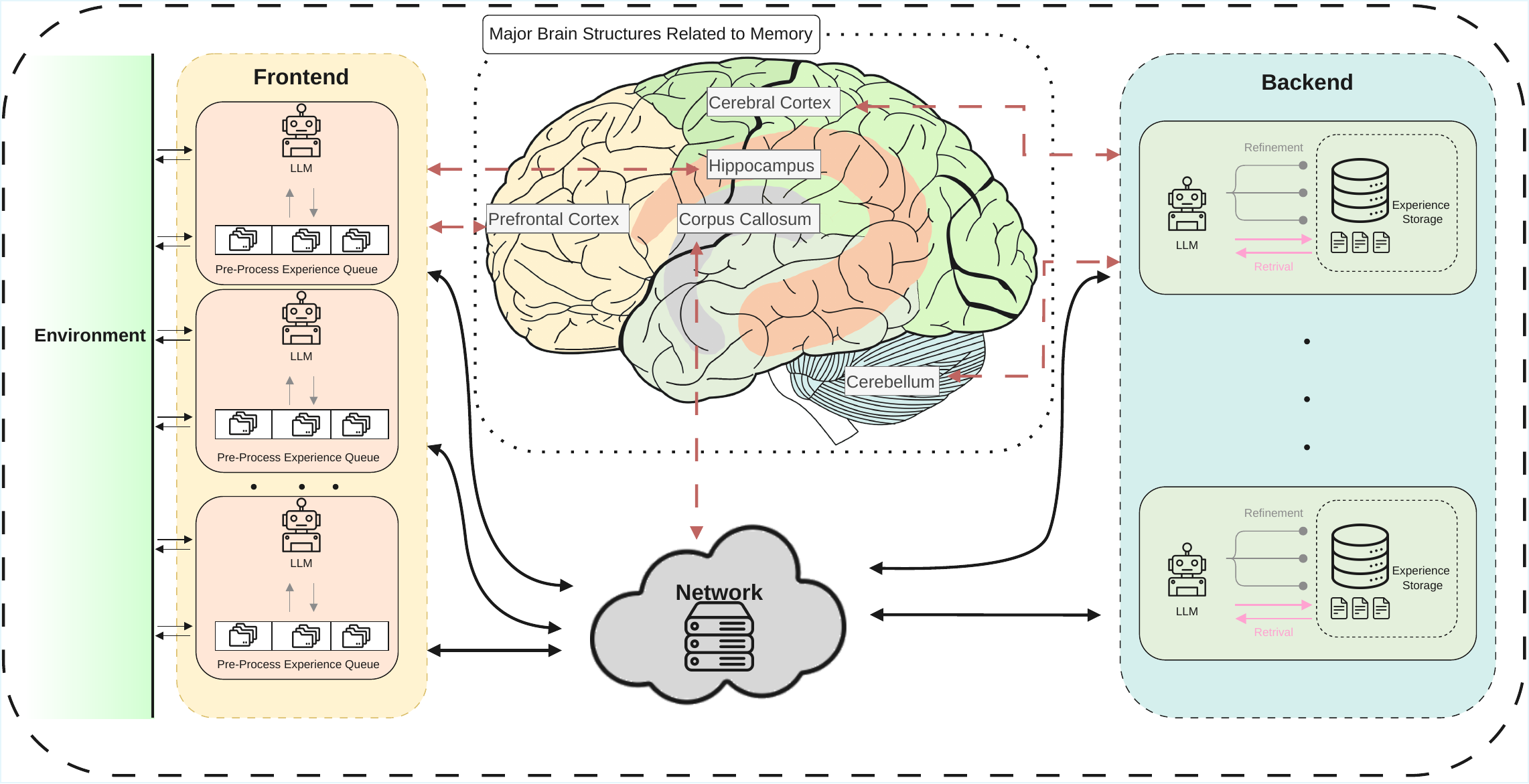}
\caption{\textbf{Architecture of Experience Scaling Framework.} The frontend, backend, and network routing components of our framework can be viewed in analogy to brain functions such as cognition, inference, memory storage, and memory transmission.}
\label{fig_framework}
\end{figure*}

This unused source represents a qualitatively new form of knowledge, capturing patterns and observations beyond human awareness or documentation. 
The heterogeneous nature of the new data introduces significant knowledge diversity in both the information and formats. Rather than storing it directly in memory, we introduce storing information in the form of experience~\cite{silver2025welcome, ijcai2024p890}.
Experience refers to the synthesized insights that LLMs refine from raw data collected from interactions with the world~\cite{10.5555/3737916.3741026}. These insights are stored in an LLM-native format optimal for the LLMs, with the purpose of enhancing future LLM inferences.
We propose shifting from scaling on static, human-generated datasets to a post-deployment evolution paradigm, termed experience scaling, that systematically captures, refines, and reuses the continuous stream of machine-acquired knowledge. By transforming deployment from a fixed endpoint into an open-ended learning process, this paradigm enables LLMs to expand their capabilities autonomously and continuously after release. To address the limitations of traditional scaling methods and implement the paradigm in practice, experience scaling rests on three core principles:

\begin{enumerate}
\item \textit{Continuous LLMs evolution after deployment.} Experience scaling enables LLMs to learn from information gathered during use, including user inputs, internal chain-of-thought traces, and multi-modal sensor data (e.g., radar, camera and thermometer)~\cite{Wu2025SkySensePlusPlus, pan2024vlp, shen2024thermometeruniversalcalibrationlarge}.  Acquired information is stored as structured experience in a dedicated memory system. Upon recei  ving a quest, relevant experience is retrieved for enhancing the quality of the outputs. This process mirrors the human mechanism of problem-solving, where prior experience similar to the situation is recalled, or external information on the same topic is sought when facing a challenge. Thus, experience scaling supports continuous, life-long evolution for LLMs, utilizing post-deployment data previously ignored by traditional scaling paradigms.

\item \textit{Acquisition of information beyond human reachability.} Informed by the human memory system~\cite{bays2024nhb}, where knowledge is not organized as discrete blocks but as an organized continuous stream of information, experience scaling processes machine-acquired data into compact, reusable representations. Direct storage of raw inputs would rapidly accumulate large volumes of redundant information after LLMs' deployment, slowing down experience retrieval efficiency when processing new quests~\cite{lee-etal-2025-shifting}. Here, the redundancy arises from semantically similar user requests and recurring patterns in sensor streams. Through distillation and condensation, our framework preserves essential information, preventing the degradation of experience quality caused by duplicated and low-value data.

\item \textit{Evolution through collaboration.} Experience scaling leverage network connection between LLMs, making contributions to a collaborative experience source. Unlike the current paradigm that improves isolated LLMs through larger datasets, more compute, or increased parameters, our framework treats environmental information as a universal asset. Experience gathered on one LLM can be transferred to others, enabling collective learning and improving performance in related scenarios across the network.
\end{enumerate}

We design the experience scaling framework with the aim of fulfilling the above core design principles. As shown in Fig.~\ref{fig_framework}, the framework follows the network-aided client-server architecture. The acquisition and storage management of experience is divided between frontend and backend, while further processing of the experience is spread throughout the experience life cycle. Our framework draws analogies to human brain functionalities~\cite{mahner2025dimensions}. The prefrontal cortex (responsible for inference) and the hippocampus (involved in information encoding) correspond to the inference and experience-generating roles of our frontend. Meanwhile, the cerebral cortex (long-term memory storage), cerebellum (memory optimization), and hippocampus (memory consolidation) have parallel functions in our backend, which handles persistent knowledge storage and refinement~\cite{moyse2024coupledneuralfieldmodel,nguyen2025cerebellar}. Furthermore, the corpus callosum, which facilitates communication between brain hemispheres, mirrors the network component that enables experience sharing across backend servers.

\begin{figure*}[h]
\centering
\includegraphics[width = 0.9\textwidth]{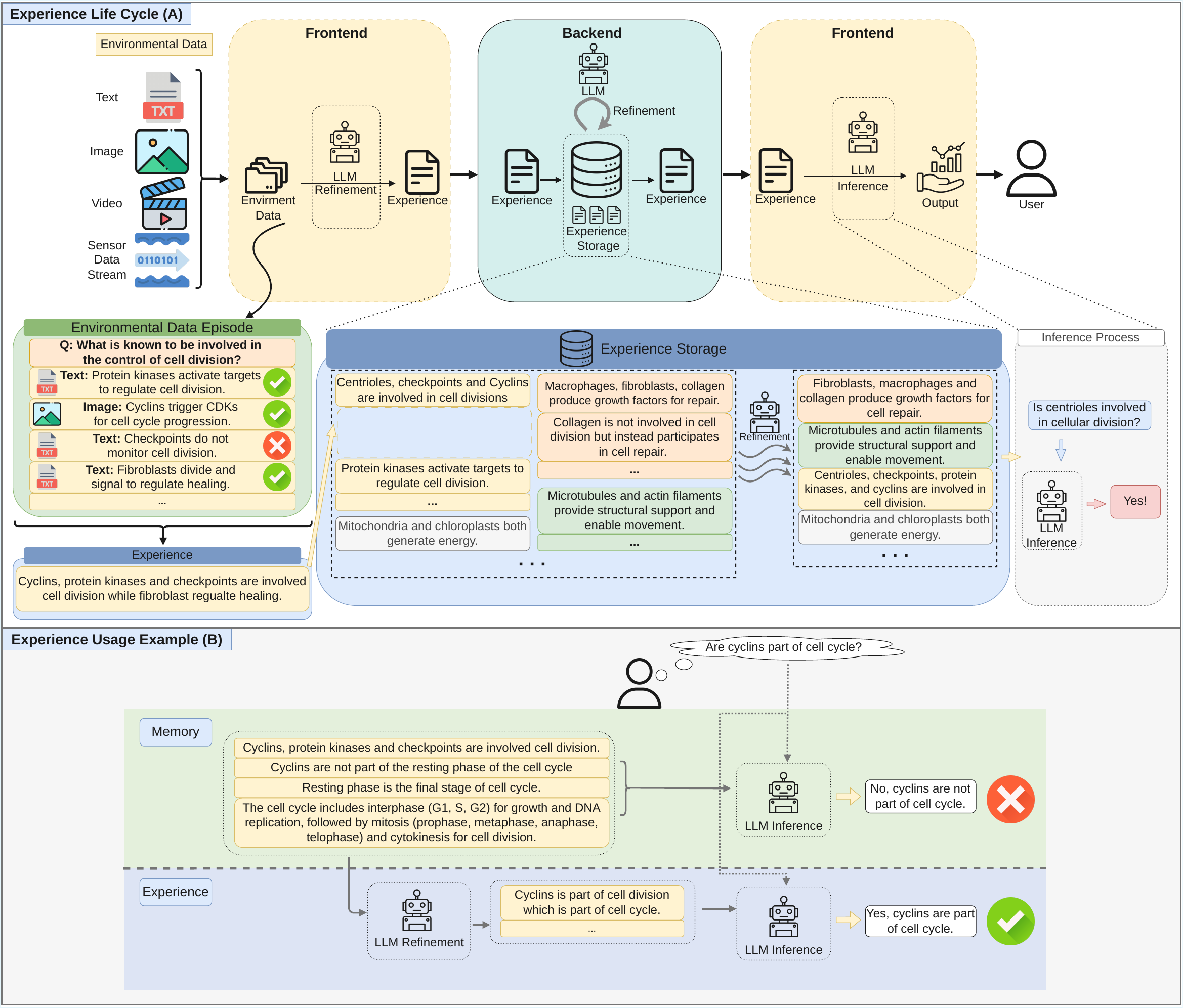}
\caption{(A) The life cycle of an experience item starting from raw environmental data processed by the frontend into experiences, stored and refined in the backend, and later retrieved to enhance LLM inference quality. (B) An example where experience refines raw information into structured insights, enabling LLMs to reason correctly beyond mere memory retrieval.}
\label{fig_lifecycle}
\end{figure*}

Figure~\ref{fig_lifecycle}(A) illustrates the detailed workflow of the framework, where the experience life cycle starts when frontend LLMs capture raw inputs from heterogeneous sources, including LLM sensor captures, user interactions, the LLMs' chain-of-thought process, and generated outputs. These inputs are temporarily aggregated and buffered into episodes. Frontend LLMs then analyze and distill episodes into a compact experience for future inference assistance. Because inputs within an episode often share high semantic similarity, this distillation step achieves substantial compression compared to the original episodic data. Experiences are subsequently transferred to backend LLMs. Storage managements aggregate these experiences by topic similarities. Optimized allocation for experience across servers reduces retrieval latency for frontend LLMs, directly improving the critical time-to-first-token (TTFT) metric \cite{Robertson2000, liu2024cachegenkvcachecompression}.

An illustrative example shown in Fig.~\ref{fig_lifecycle}(B) highlights the distinction between experience and memory. Experience emerges from the refinement of environmental information collected by LLMs. It does not merely preserve raw data but synthesizes clues into new insights and structured knowledge. Consider the case where LLMs are exposed to diverse information regarding the cell life cycle. Although the relation is not explicitly stated, the system refines its memory and derives the analysis that "cyclins are part of cell division, which in turn is part of the cell cycle." In this way, the model distills essential knowledge from environmental inputs. When asked the question "Are cyclins part of the cell cycle?", a purely memory-based retrieve may yield an incorrect inference that cyclins are not part of the cycle. In contrast, reasoning grounded in experience avoids this error, as the knowledge has already been optimized in an LLM-native form. 

Notice that the frontend and backend of experience scaling are conceptual in nature. The entire experience scaling framework can be deployed on a single machine, where the same LLM is responsible for both frontend and backend processes. In this case, the framework acts as an add-on that enables the LLM to evolve independently, with a focus on personalizing its own experience. When the framework is distributed across multiple machines functioning as frontend and backend components, it instead serves as a collective learning system that supports the joint evolution of a community of LLMs. 

\begin{figure}[h]
\centering
\includegraphics[width = 0.5\textwidth]{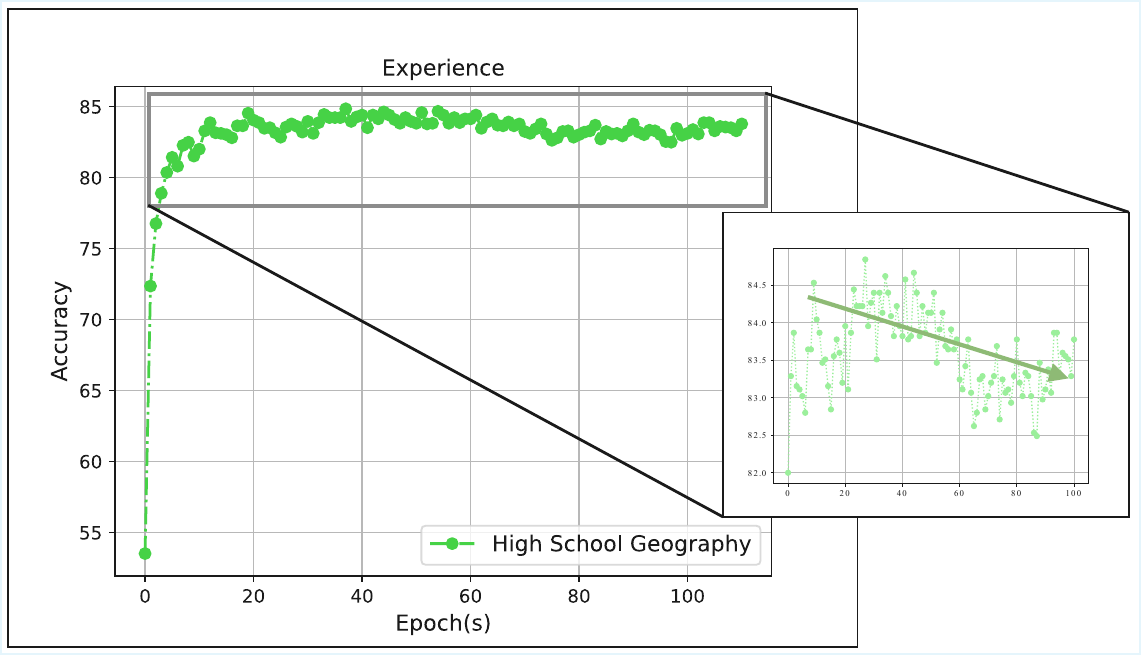}
\caption{Degradation of experience without outside interventions.}
\label{fig_too_much}
\end{figure}

Continuous accumulation of experience is not always beneficial, similar to how humans require reflection and summarization rather than only memorization~\cite{gu2024summarizingstreamdatamemoryconstrained}. We observe measurable degradation of experience quality under sustained system operation. As shown in Fig.~\ref{fig_too_much}, without external intervention, accumulated experience starts negatively impacting LLM performance over time. Motivated by this observation, we introduce server-side refinement within the experience scaling framework. The server periodically executes refinement cycles to enhance experiential knowledge quality through distillation and compression. Further details of the refinement process can be found in Section \ref{sec_method}. Finally, when receiving queries from frontend LLMs, backend LLMs retrieve relevant experiences to support improved output generation.

Overall, the proposed framework accumulates environmental information, integrates it into LLMs' world knowledge, and synthesizes it into actionable experience. It systematically utilizes the ignored potential of post-deployment experience for model enhancement. Beyond being previously unrealized information for LLM evolution, the new experience is gathered directly by LLMs through their indigenous information input channel. This new source of data stream crosses the border of human-generated data, revealing the outside world to the LLMs, which can now explore and learn autonomously. Through uncovering knowledge inaccessible to humans,  experience scaling shows potential for a new dimension for LLMs to evolve towards novel superhuman intelligence~\cite{silver2025welcome}.

\section{Results}\label{sec_results}
To validate the usefulness of continuous post-deployment knowledge build-up for the experience scaling paradigm, we designed three experiments. 
{\textit{Generalized Experience}} assesses transfer to related but unseen tasks, confirming that distilled knowledge captures reusable patterns rather than memorized answers. {\textit{Retentive Experience}} tests whether accumulated traces allow the LLM to correct errors on repeated queries. Unlike static LLMs that reproduce the same failure, experience retrieval converts past mistakes into improved future responses. {\textit{Refined Experience}} examines efficiency under saturated storage. Without refinement, redundant interactions accumulate and impair retrieval, while our server-side condensation preserves accuracy with compact representations.
Implemented with Llama-3 as both frontend and backend, the framework was evaluated on two benchmarks: MMLU, designed to assess multi-subject knowledge and reasoning, and SciQ~\cite{hendrycks2021measuringmassivemultitasklanguage, welbl2017crowdsourcingmultiplechoicescience}, a dataset of crowdsourced science examination questions.

\subsection{Generalized Experience}\label{subsec_gen_exp}

\begin{figure}[h]
\centering
\includegraphics[width = 0.5\textwidth]{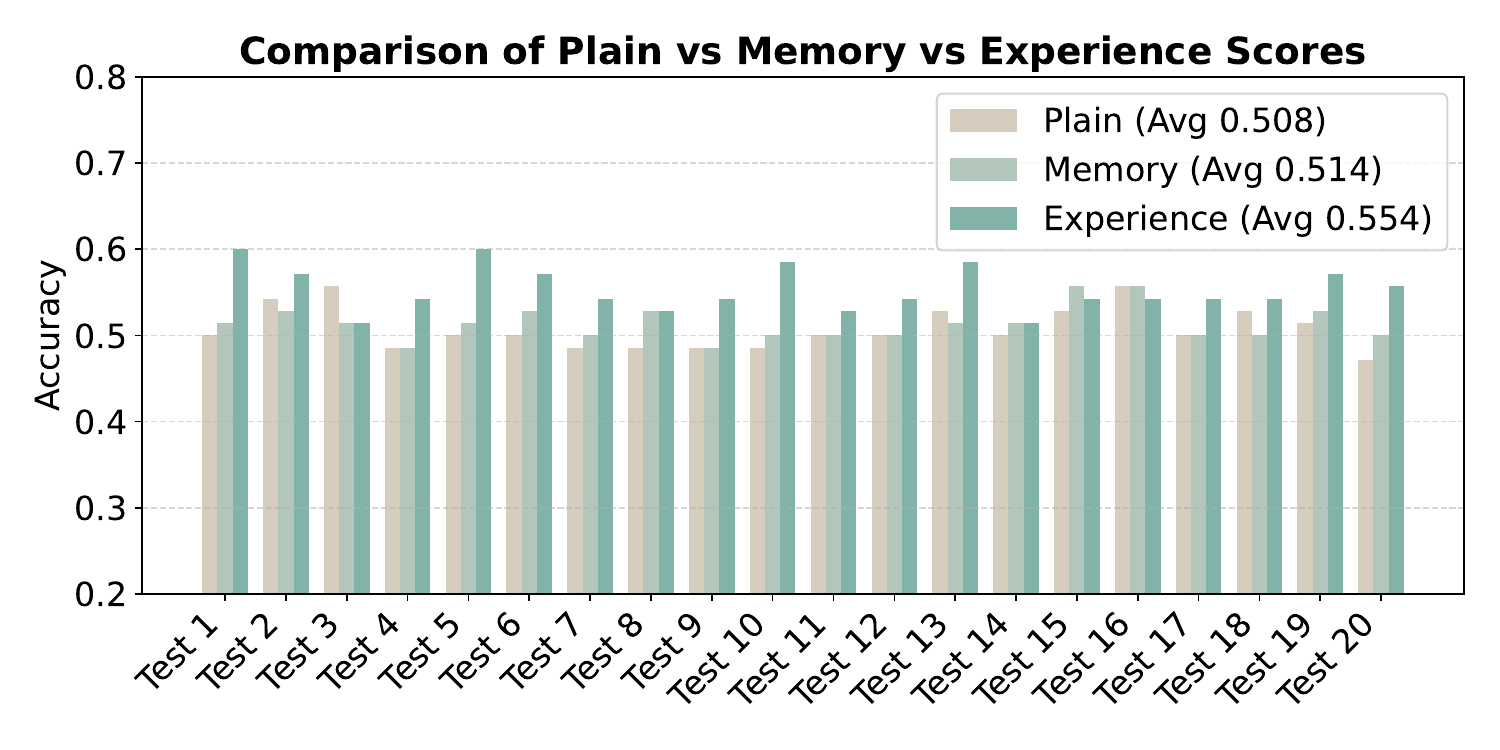}
\caption{LLMs inference performance with: plain inference, inference with memory, and inference with experience for unprecedented queries in the same topic as the memory and experience database.}
\label{fig_use_experience}
\end{figure}

This experiment demonstrates that experience-augmented LLMs generation remains robust, even when the exact query is outside the current experience database, as it allows LLMs to incorporate and leverage information from a variety of sources beyond human-generated data. The high school biology dataset from MMLU is used, where we divide the questions into a train set and a test set, each occupies 80\% and 20\% of the dataset respectively. The experience database is constructed based on the train set. 

We evaluate accuracy on the test set under three conditions: plain inference, inference with memory, and inference with experience. The results, as shown in Fig.~\ref{fig_use_experience}, demonstrate the effectiveness of experience. In addition to outperforming plain inference by 4.6\%, experience-augmented inference shows a significant accuracy improvement of 3\% over inference with memory. The experience, refined by LLMs from memory, organizes information in a way that is native to the LLM, reducing entropy noise and generating new insights from the memory. The observed difference between memory and experience supports the core principle of ``Acquisition of information beyond human reachability" in experience scaling, opening a new dimension for future improvement of LLMs.

\subsection{Repetitive Experience}\label{subsec2_rep_exp}
We evaluate experience scaling performance under a single-topic multi-user repetitive query setting. Our framework classifies tasks into different topic domains to improve output quality. Multiple clients are deployed, and each randomly samples questions from the same topic. During each epoch, every sampled question receives ten generated responses from the client. We calculate per-question accuracy as the proportion of correct responses, the derive epoch-level accuracy by averaging these values across all sampled questions within the epoch.

\begin{figure}[h]
\centering
\includegraphics[width = 0.5\textwidth]{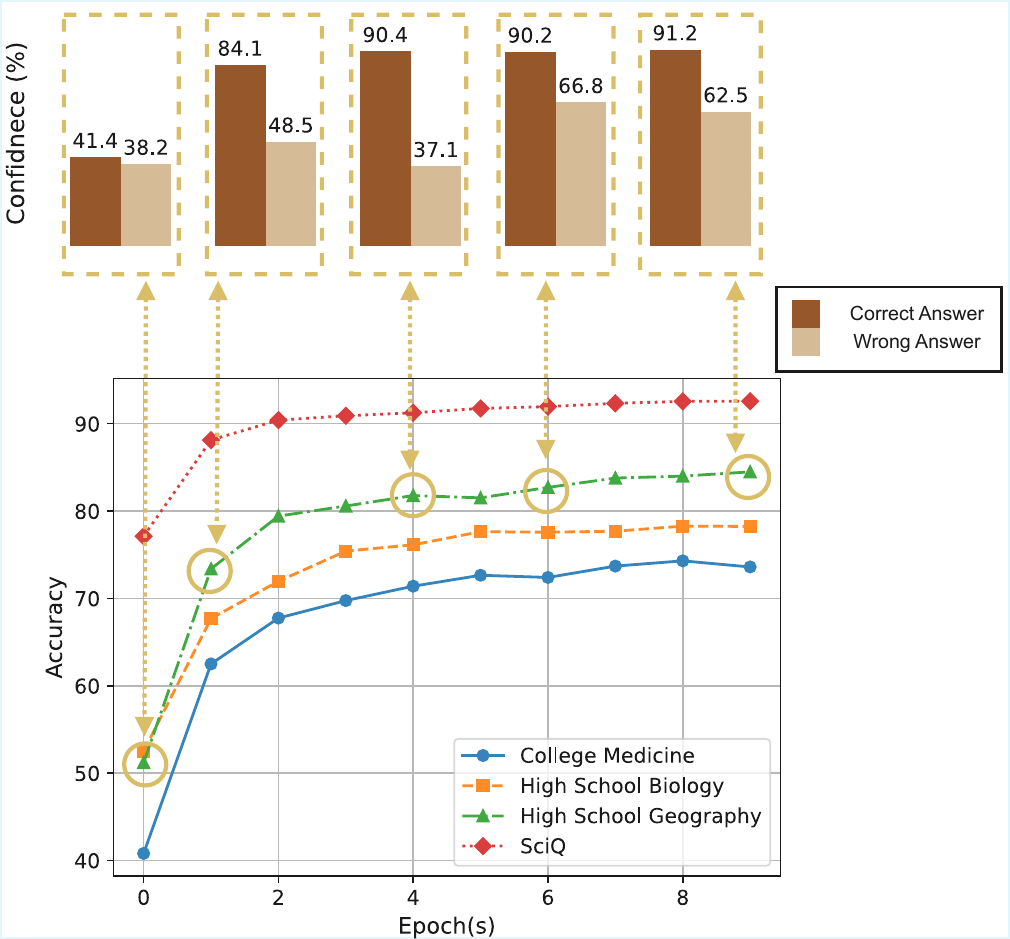}
 \caption{Top graph shows LLMs performance with the use of experience scaling. Bottom graph demonstrates the average confidence of LLMs over the output categorized by outputs' correctness.}
\label{fig_repeat_better}
\end{figure}

As shown in Fig.~\ref{fig_repeat_better}, accuracy increases steadily through epochs through College Medicine, High School Biology, and High School Geography from the MMLU dataset and the SciQ dataset. The largest relative gains appear in the earlier epochs and then gradually level off, which hits ceilings as the experience pool becomes saturated. The confidence analysis in the top panel provides additional insight. After producing an answer, the LLM is asked to report a percentage confidence in its response. These LLMs self-reported scores are consistently higher for correct predictions than for incorrect ones, demonstrating the effectiveness of experience by improving factual accuracy and by better calibrating the model’s ability to judge its own reliability.

This repetitive-query experiment illustrates how the system embodies the two core principles of experience scaling. First, it shows ``Continuous improvement after deployment" because each interaction yields distilled experience that can be stored and retrieved later, leading to consistent accuracy gains. Second, it captures ``Evolution through collaboration" since multiple clients querying the same topic contribute to and benefit from a shared backend store. Experience generated by one client directly enhances the performance of others working in the same domain. The outcome is a collaborative improvement cycle that is clearly visible in the patterns of Fig.~\ref{fig_repeat_better}. 

\subsection{Refined Experience}\label{subsec_ref_exp}
\begin{figure*}[h]
\centering
\includegraphics[width = 0.9\textwidth]{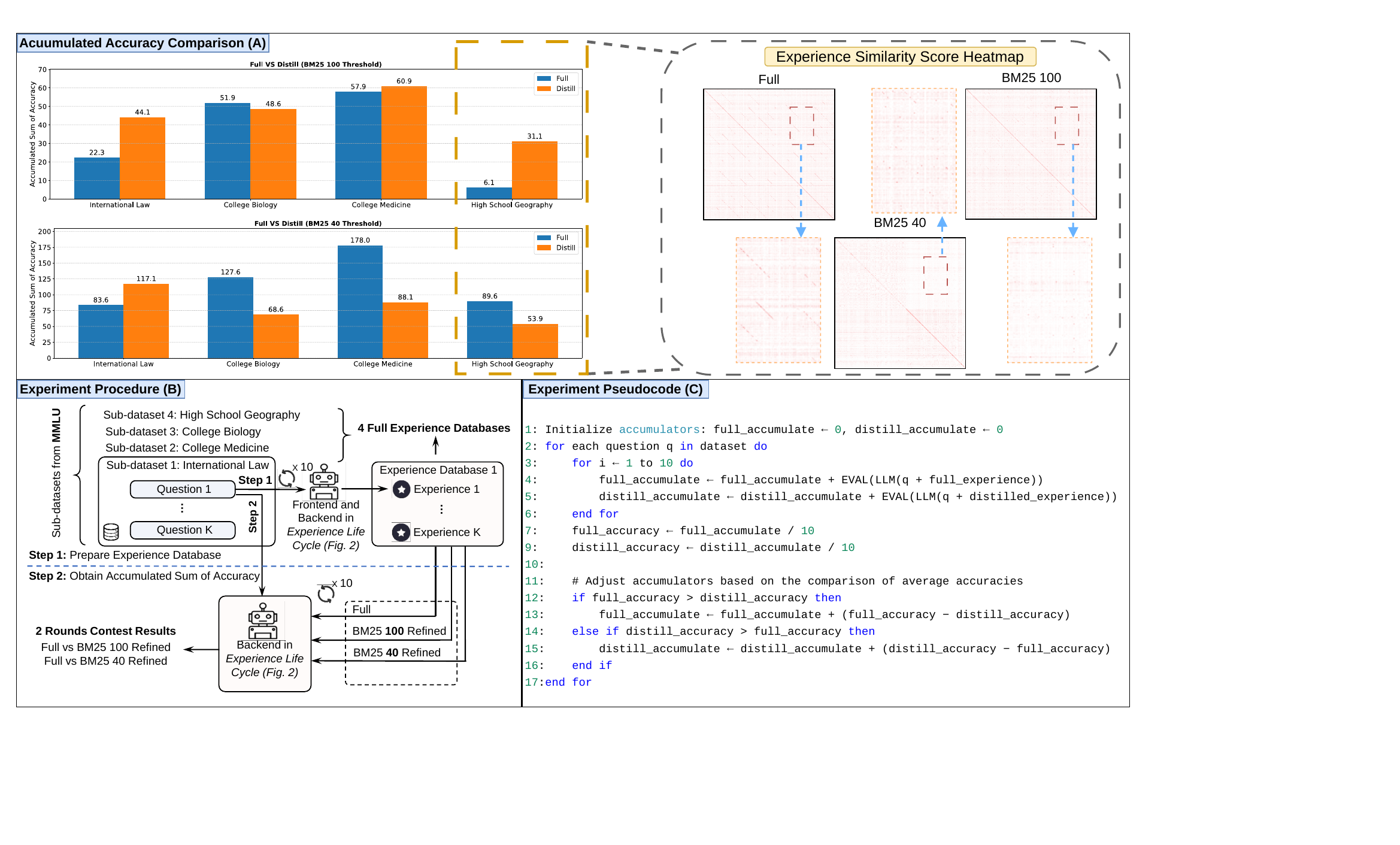}
\caption{(A) Comparison of the accumulated sum of accuracy difference between full and refined under different BM25 scores for different datasets. Similarity score heatmap of experiences for dataset ``High School Geography" under the distillation of different similarity scores. As distillation condense experiences sharing high similarity, we can see the fading of redness as the condense similarity score threshold decreases, indicating refinement of redundant experience. (B) Detailed experiment procedure for ``Refined Experience". (C) Pseudocode for the calculation process of accumulated accuracy.}
\label{fig_full_vs_refined}
\end{figure*}

We introduce experience refinement into our framework based on the observation of temporal experience degradation, illustrated in Fig.~\ref{fig_too_much}. As shown in Fig.~\ref{fig_full_vs_refined}(B), we oversaturated the experience database by executing 100 iterations per question across 4 sub-datasets from MMLU of different subjects for the experiment. Each iteration generates one experience item in the backend. We refer to the current collection of experiences stored in the backend as the full experience database. We then generate refined experience databases by applying different BM25 similarity thresholds~\cite{Robertson2000} during the condensation process. Demonstrated in the heatmaps in Fig.~\ref{fig_full_vs_refined}(A), the refinement process combines and condenses entities with similarity scores exceeding the threshold. As the BM25 threshold decreases, more entities are condensed, resulting in lower internal similarity correlation in the refined database. The detailed methodology for refinement is presented in Section~\ref{sec_method}.

Figure~\ref{fig_full_vs_refined}(C) presents the next step of the experiment, where we let LLMs generate 10 responses for each question in a dataset and obtain a question accuracy score. This process is repeated for both experience databases. We conduct a question-wise comparative analysis between each refined experience base and the full experience base, quantifying accuracy differentials and aggregating cumulative improvement metrics per configuration. We compare each refined experience base with the full experience base question-wise, calculate the difference in accuracy, and accumulate the accuracy lead for each one.

As shown in Fig.~\ref{fig_full_vs_refined}(A), LLMs assisted by a refined experience database outperform those under over-saturation conditions. It is observed that refinement with BM25 threshold of 100 consistently yields performance gains, with the maximum increase of 25\% at BM25 100 refinement occurring at the ``MMLU High School Geography'' dataset.
However, a low BM25 threshold of 40 leads to over-condensation of experiential knowledge, resulting in degraded retrieval quality due to excessive semantic compression.
One exception is the College Biology dataset, where the refined database performs worse than the full database. This phenomenon may be attributed to the domain-specific nature of college biology queries, which tend to exhibit high specificity and limited generalizability. As a result, condensation may remove critical information that is necessary for accurate responses. Table 1 offers supporting insight into the difference of datasets, showing that the College Biology dataset has significantly higher concentration similarity entities under a BM25 score of 2 (26.28\%), compared to the average of other datasets (17.81\%).

\begin{table*}[h]
\centering
\caption{Similarity Score Distribution}\label{tab_bm25}%
\begin{tabular}{@{}lllll@{}}
\toprule
\textbf{BM25 Score} & College Bio. & H.S. Geo. & College Med. & International Law  \\
\midrule
\textbf{\textless 1} & 12.18   & 10.54 & 9.28  & 4.57 \\
\textbf{1-2} & 14.1    & 11.88 & 10.61 & 6.54 \\
\textbf{2-3} & 10.77   & 11.71 & 8.61  & 8.17 \\
\textbf{3-4} & 11.56   & 12.02 & 10.13 & 9.81 \\
\textbf{4-5} & 10.61   & 11.42 & 9.31  & 9.98 \\
\textbf{5-6} & 9.13    & 9.77  & 8.61  & 9.38 \\
\textbf{6-7} & 7.76    & 7.82  & 6.94  & 8.38 \\
\textbf{7-8} & 6.05    & 6.28  & 6.45  & 7.89 \\
\textbf{8-9} & 4.58    & 4.8   & 5.27  & 6.61 \\
\textbf{9-10} & 3.33    & 3.67  & 4.55  & 5.99 \\
\textbf{10-11} & 2.48    & 2.73  & 3.69  & 4.65 \\
... & ...    & ...   & ...  & ...  \\
\bottomrule
\end{tabular}
\end{table*}

The process of refinement is integral to the post-deployment evolution of the experience scaling framework, directly reflecting on the core experience scaling principle of ``Continuous LLM Improvement after Deployment". By systematically refining accumulated knowledge, LLMs not only store data but also optimize it, enhancing both efficiency and insight. This optimization makes them more adaptable to emerging challenges. The framework mitigates the risks of overload and oversaturation through condensation and refinement, effectively distilling core information and knowledge throughout the process.

\section{Discussion}\label{sec_discussion}

As traditional scaling approaches reach diminishing returns, the next phase of LLMs evolution calls for a dynamic, experience-driven approach. By leveraging continuous, post-deployment learning, LLMs can adapt and grow autonomously, moving beyond the constraints of static data and pre-trained knowledge.

\subsection{Experience Scaling as the Second Half of Model Intelligence Growth}
The results show that the proposed paradigm is able to embody the experience scaling framework, fulfilling the three core principles for extending scaling into a new dimension. 
A crucial implication is that this shift represents the necessary ``second half"  of model intelligence growth. 
The first half has been driven by scaling datasets, parameters, and computing resources, which enabled LLMs to tackle increasingly complex problems, but is now reaching a natural ceiling as high-quality human-generated data becomes exhausted. The second half can be viewed as a stage where LLMs must seek growth beyond human-created data, and experience scaling represents one concrete form of this stage. By systematically accumulating, distilling, and reusing these experiences, LLMs can move beyond static pretraining and the inherent limits of human data. This transition opens the possibility of capability growth that surpasses what human-generated content alone can provide, positioning experience scaling as a necessary step toward intelligence beyond human boundaries.

\subsection{Distinction from Continual Learning and Memory Systems}
While experience scaling may appear superficially similar to traditional continual learning and LLM memory systems, its framework and objectives are fundamentally different~\cite{freire2025sequential, lewis2021retrievalaugmentedgenerationknowledgeintensivenlp}. 
Unlike continual learning, which focuses on incrementally updating model weights, experience scaling avoids modifying the base LLM parameters~\cite{hagele2024scaling}. It constructs an external, collaboratively refined knowledge repository that evolves independently, preventing catastrophic forgetting and preserving the stability of the core model~\cite{Kirkpatrick_2017, zheng2025learning}.
Experiences are stored in a structured format and undergo periodic refinement. During inference, relevant experiences are retrieved to augment the LLMs outputs, creating a closed-loop system of collection, distillation, refinement, and reuse.
Memory systems primarily focus on storage, but experience scaling extends further by actively managing, refining, and allocating stored knowledge across distributed servers with optimized strategies. 
Crucially, experience scaling shifts the adaptation from a model-centric process to environment-driven knowledge synthesis, enabling LLMs to autonomously discover and refine patterns beyond human-generated data. This shift opens up possibilities for capabilities that cannot be attained through static pretraining or simple memory retrieval.

\subsection{Benefits of the Experience Scaling Paradigm}
Our experience scaling framework has several notable benefits embodying the three core principles of the experience scaling paradigm. 
First, it breaks the ceiling of traditional scaling. Current models trained with larger datasets and parameters face diminishing returns, with performance gains slowing while computational and energy costs continue to rise~\cite{porian2024resolving}. By enabling continuous post-deployment evolution, experience scaling offers a new growth dimension that is not bound by the limitations of pre-training, allowing LLMs to improve in efficiency and capability without exponential resource demands. Second, it enhances long-term adaptability and personalization. Through systematic accumulation and refinement of environmental experience, LLMs in an experience scaling framework can rapidly incorporate new knowledge, adjust to domain-specific requirements, and sustain relevance in dynamic contexts such as healthcare diagnostics, autonomous systems, or financial risk monitoring, where static pre-trained knowledge quickly becomes outdated. Third, it establishes a foundation for collective intelligence across LLMs. Experience scaling transforms knowledge from being locked within isolated LLMs into a transferable asset that can be shared across networked systems. This collaborative learning mechanism enables faster dissemination of task-relevant experience, reduces redundancy in model training, and accelerates collective progress, much like scientific communities thrive through shared knowledge rather than isolated discovery.

\subsection{Limitations and Future Directions}
The current framework exhibits several limitations that present opportunities for future enhancement to better advance the experience scaling paradigm. 

\begin{enumerate}
\item \textit{Client-Server model.} While the client-server design provides a clear separation of LLM functionality, the associated communication latency introduces overhead on inference tasks~\cite{Gu_2021}. Furthermore, dedicating backend LLMs leads to significant underutilization of compute resources during idle periods. Future work can explore a serverless peer-to-peer framework for experience scaling, improving resource efficiency, and reducing retrieval latency.

\item \textit{Multi-Modality.} One of the core design principles of experience scaling is to encompass LLMs inputs of different forms. This includes user text, uploaded images, and videos captured by LLMs sensors. The current framework centers on textual experiences, but future work can extend it to multi-modal domains to achieve broader applicability.

\item \textit{LLM-Native Experience.} The current memory system in the experience scaling framework stores experience in the form they were originally input into the frontend. The human-readable experience is easy to analyze. However, we believe that not storing experience in LLM-native form would suffer loss of information~\cite{maharana-etal-2024-evaluating, Tiezzi2025BackTR}. Similar to how semantic richness can be lost in the translation of literature between languages, preserving experience in an LLM-native representation may retain deeper structures and enable more effective optimization.

\end{enumerate}

\section{Methods}\label{sec_method}
Experience scaling, as illustrated in Fig.~\ref{fig_framework}, gathers observations richly grounded in the environment. In this section, we outline the frontend and backend of experience scaling, dive into details of the creation, storage operation and utilization of experience.

\subsection{Experience Scaling Client}\label{subsec_client}
Clients manage direct interactions with the environment and generate experience.
In this framework, users and environmental sensors interface only with a client that hosts frontend LLMs. At the end of each interaction interval, all environmental data generated during that period, whether from the environment or generated by LLMs, are stored as an episode. These episodes enter a client-managed processing queue where raw unstructured environmental information is transformed into experiences. For optimizing user experience, client LLMs prioritize real-time inference for user queries. When idle, they process queued episodes and synthesize experiences through condensation and summarization algorithms.

A core feature of experience scaling is the framework-wide distribution of experiential knowledge. To better utilize diverse sources of inputs, newly processed experiences are categorized by content types. These classified experiences are then routed to target servers through universal routing. When receiving a query, client first determines the type of the task, and then retrieves relevant experience for better optimized generation.

\subsection{Experience Scaling Server}\label{subsec_server}

Experience scaling servers refine the knowledge database, as shown in Fig.~\ref{fig_lifecycle}. The process condenses stored experience once the database reaches a predefined threshold, reducing the risk of oversaturation. The database condensation process begins with the server computing a pairwise similarity score matrix $S$ across all documents as

\begin{equation}
S_{ij}=\mathrm{Similarity\_Score}(\mathbf{w}_i,\mathbf{w}_i),
\label{smatrix}
\end{equation}

\noindent where $W = \{w_1, w_2, \ldots, w_n\}$ denotes the experience database of size $n$, with $w_i$ and $w_j$ representing the $i$-th and $j$-th experience respectively. The resulting matrix $S \in \mathbb{R}^{n \times n}$ contains the pairwise similarity scores between all experiences.

All experiences are initially placed in the unused group $U$. 
High-similarity neighbors of an anchor $i$ are grouped as

\begin{equation}
G_i=\{j \in U \mid S_{ij}\ge \tau_{\mathrm{sim}}\,\},
\label{group}
\end{equation}

\noindent where $\tau_{\mathrm{sim}}$ is the similarity threshold used for condensation.

Each high-similarity group $G$ is then condensed by an LLM into a single experience

\begin{equation}
c_G=\Phi\!\left(\{w_j\}_{j\in G}\right)
=LLM_{{\textrm {Condense}}}\!\left(\bigoplus_{j\in G} w_j\right),
\label{eq_groupcondense}
\end{equation}

\noindent where $\Phi$ denotes the condensation operator and $\bigoplus$ indicates content aggregation.

Finally, the database is updated by replacing grouped items with their condensed representatives:

\begin{equation}
W'=\{\,w_u:\;u\in U\setminus (\cup_i G_i)\,\}\;\cup\;\{\,c_G:\;G\in \mathcal{G}\,\},
\label{newbase}
\end{equation}

\noindent where $\mathcal{G}$ is the set of non-empty groups produced by (\ref{eq_groupcondense}). This condensation reduces redundancy and controls database growth.

Servers store experience partitioned by types. During query retrieval processing, each server retrieves the top-k most relevant experiences using BM25 similarity scoring against the query. The similarity score ranking function we employ for both the refinement process and the retrieval processes is Okapi BM25~\cite{Robertson1994OkapiAT}:

\begin{equation}
\text{BM25}(q, d) =  \frac{\sum_{t \in q} {\textrm {IDF}}(t) \cdot f(t, d) \cdot (k_1 + 1)}{f(t, d) + k_1 \cdot \left(1 - b + b \cdot \frac{|d|}{\text{avgdl}}\right)},
\label{bm25}
\end{equation}

\noindent where in the retrieval stage, we first quantify the lexical relevance between a user query $q$ with every candidate document $d$ in the experience database. 

BM25 aggregates, over query terms $t$, a product of two factors: an inverse document frequency ${\textrm {IDF}}(t)$

\begin{equation}
{\textrm {IDF}}(t) = \log \frac{N - n_t + 0.5}{n_t + 0.5}.
\label{idf}
\end{equation}

\noindent which captures how informative $t$ is across the corpus, and a term-frequency component that increases with the count $f(t,d)$ but saturates under the control of hyperparameter $k_1$. 
$N$ in (\ref{idf}) is the total number of documents, and $n_t$ is the number containing $t$. The {\textrm {IDF}} score is inversely proportional to term frequency $t$, diminishing the weight of common, uninformative words (e.g., ``the", ``is") in the final similarity calculation.

The denominator in BM25 calculation applies length normalization so that long documents do not dominate purely by containing more tokens, with strength governed by $b$ via the ratio of the document length $|d|$ to the corpus average $\text{avgdl}$.

Following BM25 computation, experiences scoring below a return threshold are discarded to ensure retrieval relevance. The remaining experiences are sorted in descending order by similarity score and lexicographically by text to ensure deterministic ordering. Finally, this sorted list is truncated up to the top-$K$ experiences before being returned to the client.

\subsection{Experience Scaling Network}\label{subsec_network}
After the client generates an experience from environmental data, it classifies the experience into different types based on its content. As demonstrated in Fig.~\ref{fig_framework}, the experience is routed to the appropriate server for storage and refinement once categorized. This approach improves collaboration across LLMs by ensuring that experiences are shared within the correct domain, aligning with the core principle of ``Evolution through collaboration." It also enables efficient access to specialized knowledge, thereby improving overall system performance.

When a new query is received, the client first determines its type and then routes it to the relevant server. The server then retrieves relevant experiences from its experience database. This targeted retrieval process improves both efficiency and accuracy.

\section{Conclusion}
Experience scaling introduces a new dimension for advancing LLMs beyond the limits of static human-generated datasets. By enabling continuous post-deployment evolution through autonomous interaction with the environment, experience scaling transforms repeated interactions into improved responses, sustains efficiency under long-term accumulation, and supports generalization to related but unseen tasks. Collaborative sharing of distilled experience further amplifies these benefits across deployed LLMs. This paradigm directly addresses the data saturation and diminishing returns of traditional scaling laws, offering a path for LLMs to evolve after release. We expect these findings to motivate future research and further drive LLMs development towards the next stage of intelligence.



\bibliographystyle{IEEEtran}
\bibliography{Ref}
\end{document}